\documentclass[11pt,a4paper]{article}
\usepackage[hyperref]{acl2021}
\usepackage{times}
\usepackage{latexsym}

\usepackage{microtype}
\usepackage{times}
\usepackage{latexsym}
\usepackage{amsmath}
\usepackage{amsthm}
\usepackage{amsfonts}
\usepackage{booktabs}
\usepackage{bbm}
\usepackage{graphicx}
\graphicspath{ {Figure/} }
\usepackage{url}
\usepackage{xcolor}
\usepackage{multirow}
\usepackage{algorithm}
\usepackage{algorithmic}

\aclfinalcopy 

\title{LAWDR: Language-Agnostic Weighted Document Representations from Pre-trained Models}

\author{Hongyu Gong, Vishrav Chaudhary, Yuqing Tang and Francisco Guzm\'{a}n \\
  Facebook AI \\
  \texttt{\{hygong, vishrav, yuqtang, fguzman\}@fb.com} 
 }

\date{}

\begin{document}
\maketitle
\begin{abstract}
Cross-lingual document representations enable language understanding in multilingual contexts and allow transfer learning from high-resource to low-resource languages at the document level. Recently large pre-trained language models such as BERT, XLM and XLM-RoBERTa have achieved great success when fine-tuned on sentence-level downstream tasks. It is tempting to apply these cross-lingual models to document representation learning. However, there are two challenges: (1) these models impose high costs on long document processing and thus many of them have strict length limit; (2) model fine-tuning requires extra data and computational resources, which is not practical in resource-limited settings. In this work, we address these challenges by proposing unsupervised Language-Agnostic Weighted Document Representations (LAWDR). We study the geometry of pre-trained sentence embeddings and leverage it to derive document representations without fine-tuning. Evaluated on cross-lingual document alignment, LAWDR demonstrates comparable performance to state-of-the-art models on benchmark datasets.
\end{abstract}

\section{Introduction}
Language representations map words, sentences or documents to dense vectors, capturing their semantic meaning \cite{alpolyglot}. Cross-lingual representations project different languages into the same vector space.
Recently there has been a surge of research on pre-training cross-lingual language models with large-scale corpora. These studies include LASER \cite{artetxe2018massively}, multilingual BERT \cite{devlin2019bert}, XLM \cite{lample2019cross} and XLM-RoBERTa \cite{conneau2019unsupervised}. 
Multilingual sentence representations produced by these models have led to strong improvements in downstream sentence-level tasks. 

Despite the extensive study of cross-lingual sentence representations, there has been few works on cross-lingual document embeddings. Transformer based models such as BERT, XLM and XLM-RoBERTa have strict length limits on inputs and therefore cannot process long documents. While complex, neural-based models have been proposed, averaging sentence embeddings has remained a hard-to-beat baseline 
\cite{guo2019hierarchical}. 
Here, we propose a document-level model based on weighted sentence embeddings, and the sentence weights are inversely related with its density in the corpus, akin to an inverse corpus frequency.\footnote{The intuition behind sentence weighting is that a sentence with higher density in a corpus is less informative and unique to its document, and thus should be assigned with a lower weight.} 

Cross-lingual representations 
enable the use of the same model for downstream tasks, 
regardless of their languages. In this work, we reveal that pre-trained cross-lingual embeddings are not language-agnostic, and that the language bias hurts their semantic representation power. We propose to remove the language bias efficiently and derive language-agnostic sentence representations. We show that debiased representations improve performance on cross-lingual tasks without the need for fine-tuning.



In this work, we propose Language-Agnostic Weighted Document Representation (LAWDR). Our document representations are built upon sentence embeddings obtained from off-the-shelf pre-trained language models. We first obtain language-agnostic sentence embedding by removing their language bias. Then we derive document representations from a weighted combination of these sentence embeddings. LAWDR is an unsupervised approach  without model fine-tuning and task-specific supervision.  

Cross-lingual document representations can be naturally applied to align multi-lingual document alignment and mine parallel data for machine translation \cite{buck2016findings}.
To measure document similarity, we study and compare two similarity metrics, cosine similarity \cite{mikolov2013distributed} and the margin function \cite{artetxe2018margin}. Empirically, the margin metric is a better measure of cross-lingual document similarity.
The learned document representation is evaluated on two public datasets from WMT16 and WMT19 shared tasks, and achieves strong performance in cross-lingual alignment. 




We summarize our contributions below: \\
\noindent(1) We reveal that state-of-the-art pre-trained models carry language bias which hurts their semantic representation power. This work proposes an efficient approach to derive language-agnostic sentence embeddings.\\
\noindent(2) We propose an unsupervised approach LAWDR to derive language-agnostic document representations from sentence embeddings. It achieves strong empirical performance in cross-lingual document alignment. 


\section{Language Bias in Sentence Embeddings}
Recently state-of-the-art performance has been achieved in cross-lingual tasks by pre-trained models such as BERT \cite{devlin2019bert}, XLM \cite{lample2019cross} and XLM-RoBERTa (abbreviated as XLM-R in the following discussions) \cite{conneau2019unsupervised}. These models learn contextualized representations for tokens in given input sentences. There are two commonly used approaches to obtain sentence representations from the token embeddings. One approach is to concatenate special tokens such as \text{[CLS]} in BERT, $\langle\backslash s\rangle$ in XLM and $\langle s\rangle$ in XLM-R with a sentence as the input sequence to the pre-trained models. Embeddings of these special tokens are taken as the sentence embedding \cite{devlin2019bert,lample2019cross}. We refer to this method as special token (ST) approach for simplicity, and it demonstrates good performance on sentence-level tasks when fine-tuned on external labeled datasets. 

The other way is mean pooling (MP), deriving sentence embeddings as the average of token embeddings. It is a simple yet efficient approach to encode compositional word semantics into sentences representations \cite{le2014distributed}. In this work, we study and compare sentence embeddings derived using these two approaches.

\begin{figure*}[h]
\centering
\begin{minipage}{0.32\textwidth}
\centerline{\includegraphics[width=\linewidth]{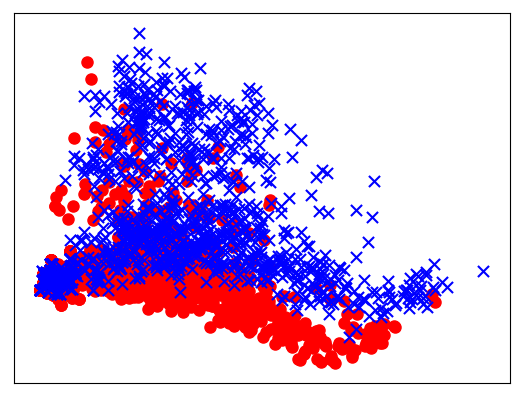}}
\centerline{\small{(a) BERT embeddings}}
\label{fig:laser_bias}
\end{minipage}
\begin{minipage}[c]{0.32\textwidth}
\centerline{\includegraphics[width=\linewidth]{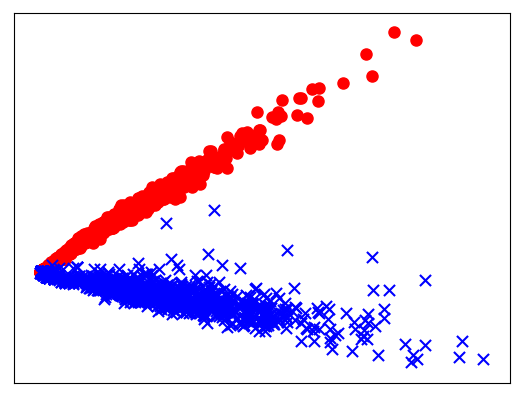}}
\centerline{\small{(b) XLM embeddings}}
\label{fig:bert_bias}
\end{minipage}
\begin{minipage}[c]{0.32\textwidth}
\centerline{\includegraphics[width=\linewidth]{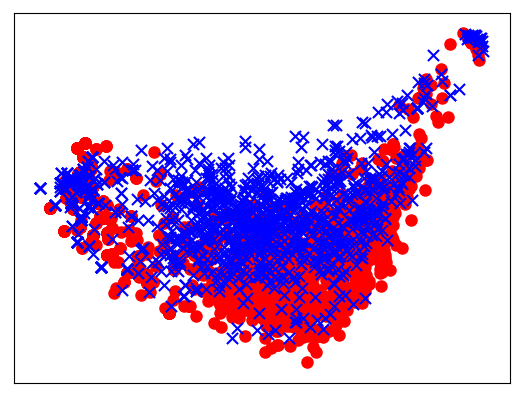}}
\centerline{\small{(c) XLM-R embeddings}}
\label{fig:xlm_bias}
\end{minipage}
\caption{Visualization of MP sentence embeddings by pre-trained models. English sentences are marked as red and French sentences are blue.}
\label{fig:bias_visualization}
\end{figure*}

Cross-lingual representations are expected to be language-agnostic. However, we argue that this is not the case, as we will show both qualitatively and quantitatively that the pre-trained sentence embedding carries language signals. 

\noindent\textbf{Qualitative analysis}. We extract parallel documents in English and French collected from the domain of eu2007.de in the WMT-16 document alignment dataset \cite{buck2016findings}. These parallel documents consist of $1,356$ English and $1,295$ French sentences. We project their sentence embeddings to two-dimension space using principal component analysis (PCA). Fig.~\ref{fig:bias_visualization} visualizes MP sentence embeddings derived from pre-trained BERT, XLM and XLM-R. It shows that sentences are separated in the vector space based on their languages even if they are from parallel documents. ST sentence embeddings have similar distribution as MP, which is shown in the supplementary material.

\begin{table}[htbp!]
\centering
\begin{tabular}{cccccc}
\hline
Embedding & BERT & XLM & XLM-R \\ \hline
$\{\mathbf{v}_{s}\}_{s}$  & 95.3 & 97.3 & 94.7 \\ 
$\{\tilde{\mathbf{v}_{s}}^{\text{lang}}\}_{s}$ & 100.0 & 100.0 & 97.6 \\ 
$\{\tilde{\mathbf{v}_{s}}^{\text{sem}}\}_{s}$ &  52.5 & 53.5 & 52.0 \\ \hline
\end{tabular}
\caption{Language classification accuracy of MP embeddings. ($\mathbf{v}_{s}$: sentence embedding from a pre-trained model, $\tilde{\mathbf{v}_{s}}^{\text{lang}}$: language component and $\tilde{\mathbf{v}_{s}}^{\text{sem}}$: semantic component estimated from sentence embeddings.)}
\label{tab:lang_clf}
\end{table}

\noindent\textbf{Quantitative analysis}. To further justify our conjecture that pre-trained sentence embeddings carry language bias, we design a language classification task on the same data. Parallel documents are randomly split into train and test sets with a ratio of 8:2. A linear Support Vector Machine (SVM) is trained with MP sentence embeddings to classify the language of the given sentences. In the row of $\{\mathbf{v}_{s}\}_{s}$, Table~\ref{tab:lang_clf} 
reports the language classification accuracy on test sentences. Sentence languages could be classified with high accuracy above $90\%$ for all pre-trained models. ST sentence embeddings have similar results, which are reported in the supplementary material. This quantitatively demonstrates that language signals are encoded to these pre-trained representations.

\section{Language-Agnostic Sentence Embeddings}
We have shown that the sentence embeddings from pre-trained models are biased towards languages. Hence the cross-lingual representations are actually a mixture of language and semantic signals. Suppose that the sentence $s$ has an embedding $\mathbf{v}_{s}$ in the $r$-dimension space. We assume that the embedding space can be decomposed into two orthogonal subspaces: a language subspace $L$ and a semantic subspace $S$. Suppose that the basis vectors of space $L$ is $\{\mathbf{u}^{\text{lang}}_{i}\}_{i=1}^{m}$, and the basis of space $S$ is $\{\mathbf{u}^{\text{sem}}_{j}\}_{j=1}^{n}$. The basis vectors are orthogonal to each other within a subspace. Moreover, basis vectors from the two subspaces are also orthogonal, i.e., $\langle \mathbf{u}^{\text{lang}}_{i},\mathbf{u}^{\text{sem}}_{j}\rangle=0$, considering that subspace $L$ is orthogonal to $S$. Therefore, the sentence embedding $\mathbf{v}_{s}$ can be expressed as a sum of its language component $\mathbf{v}_{s}^{\text{lang}}$ and semantic component  $\mathbf{v}_{s}^{\text{sem}}$.  
\begin{align}
    \label{eq:decompose_vector}
    \mathbf{v}_{s} = \mathbf{v}^{\text{lang}}_{s} + \mathbf{v}^{\text{sem}}_{s}.
\end{align}
The language component can be represented as a linear combination of language basis vectors, i.e., \(\mathbf{v}^{\text{lang}}_{s}=\sum_{i=1}^{m}\alpha_{s,i}\mathbf{u}^{\text{lang}}_{i}\), where the coefficient $\alpha_{s,i}$ is its projection onto the basis $\mathbf{u}^{\text{lang}}_{i}$. 
Similarly, the semantic component consists of semantic basis vectors $\mathbf{v}^{\text{sem}}_{s}=\sum_{j=1}^{n}\beta_{s,j}\mathbf{u}^{\text{sem}}_{j}$, and $\beta_{s,j}$ is the projection onto $\mathbf{u}^{\text{sem}}_{j}$.

\subsection{Language Classification}
Consider two sentences  $s_{1}$ and $s_{2}$ with embeddings $\mathbf{v}_{s_{1}}$ and $\mathbf{v}_{s_{2}}$ respectively. We now explain the phenomenon that embeddings could be classified by their languages with the assumption that the embedding is a mixture of language and semantic components.

Given the orthogonality of language and semantic spaces, we can write the inner product between the sentence embeddings in Eq.~(\ref{eq:decompose_vector}) as below.
\begin{align}
\label{eq:inner_prod}
\langle \mathbf{v}_{s_{1}},\mathbf{v}_{s_{2}}\rangle 
= \langle \mathbf{v}^{\text{lang}}_{s_{1}}, \mathbf{v}^{\text{lang}}_{s_{2}}\rangle + \langle \mathbf{v}^{\text{sem}}_{s_{1}}, \mathbf{v}^{\text{sem}}_{s_{2}}\rangle
\end{align}

Eq.~(\ref{eq:inner_prod}) shows that the high correlation of language components leads to large inner product between embeddings for sentences in the same language. The sentence embeddings from the same language stay close in the vector space. As a result, the embeddings are linearly separable according to their languages. 

\subsection{Debiasing Sentence Embeddings}

Based on Eq.~(\ref{eq:inner_prod}), language components affect the inner product of sentence embeddings, and limit the semantic representation power of pre-trained embeddings.
An immediate question is how to derive language-agnostic sentence embeddings.

Previous works rely on linguistic resources such as bilingual dictionaries to transform embeddings in different languages to the same space \cite{schuster2019cross,conneau2020emerging}. We propose an unsupervised debiasing method, which removes language components and leaves language-agnostic representations without using external resources. 

Since language components are highly correlated for sentences in the same language, language bias is the common components existing in their embeddings. We decompose sentence embeddings in the same language using Singular Value Decomposition (SVD), and identify the dominant components which capture the largest variance of these embeddings as the language components. Given embeddings $\{\mathbf{v}_s: s\in C_{l}\}$ of a set of sentences $C$ in language $l$, the dominant components $\{\tilde{\mathbf{u}}^{\text{lang}}_{i}\}_{i=1}^{m}$ of these sentence embeddings is derived as the estimated basis of the language subspace.

With the identified language subspace, we could estimate the language component $\tilde{\mathbf{v}}_{s}^{\text{lang}}$ of the sentence embedding $\mathbf{v}_{s}$: $\tilde{\mathbf{v}}_{s}^{\text{lang}}=\sum\limits_{i=1}^{m}\tilde{\alpha}_{s,i}\tilde{\mathbf{u}}^{\text{lang}}_{i}$. The weight coefficient $\tilde{\alpha}_{s,i}$ is the projection of $\mathbf{v}_{s}$ onto the basis of language subspace $L$, i.e.,  $\tilde{\alpha}_{s,i}=\langle \mathbf{v}_{s}, \tilde{\mathbf{u}}_{i}^{\text{lang}} \rangle$.

We debias the sentence embedding $\mathbf{v}_{s}$ by removing its language component and approximate the semantic representation $\tilde{\mathbf{v}}^{\text{sem}}_{s}=\mathbf{v}_{s} - \tilde{\mathbf{v}}_{s}^{\text{lang}}$ .

In our experiments, the rank $m$ of language subspace $L$ is a hyperparamter discussed in the supplementary material.

To prove the effectiveness of this debiasing method, we designed an experiment in which only partial information is available to train a language classifier. We reuse the same set of sentences in the language classification task. The linear SVM classifier takes as inputs the language component $\{\tilde{\mathbf{v}}_{s}^{\text{lang}}\}_{s}$ and the semantic components $\{\tilde{\mathbf{v}}_{s}^{\text{sem}}\}_{s}$ respectively. The test accuracy of language classification is reported in the last two rows of Table~\ref{tab:lang_clf}.

For all models, language components achieve almost perfect language classification while the performance of semantic components is as poor as  random guess. Hence we empirically demonstrate that language signals exist in the dominant components of sentence embeddings. Moreover, the language bias is mitigated in the debiased embedding $\tilde{\mathbf{v}}^{\text{sem}}_{s}$. 
In the following study of document representation, we will denote the debiased sentence embedding $\tilde{\mathbf{v}}^{\text{sem}}_{s}$ as $\tilde{\mathbf{v}}_{s}$ for simplicity, and use it as the language-agnostic sentence representation.

\section{Document Representation Model}
\label{sec:doc_emb}

In this section, we will discuss how to derive language-agnostic document representation from the debiased sentence embeddings. 

\subsection{Weighted Document Representation}

Averaging is a simple yet strong approach to represent compositional semantics. Strong performance comparable to supervised neural models has been achieved by sentence embeddings derived from average word embeddings \cite{arora2016simple} and document embeddings from average sentence embeddings \cite{guo2019hierarchical} respectively.

As an improvement over the unweighted average, recent works learn sentence embeddings as a weighted average of word embeddings, where the weight is negatively correlated with the word frequency \cite{arora2016simple,ethayarajh2019contextual}. Their intuition is that frequent words such as stop words carry less semantic information, and that they should contribute less to sentence embeddings.

Borrowing their ideas, we propose to learn document representation from a weighted linear combination of the debiased sentence embeddings. Different from words, sentences have continuous probability distribution. Hence we use kernel density estimator (KDE) to efficiently estimate sentence density from sentence embeddings \cite{scott2015multivariate}

Considering that sentence with low density could better discriminate documents, we use the inverse density of sentences as their weights. We derive the document representation $\mathbf{v}_{d}$ below
\begin{align}
\mathbf{v}_{d} = \sum\limits_{s\in d}w_{s}\tilde{\mathbf{v}}_{s},
\end{align}
where the weighting coefficient $w_{s}$ of sentence $s$ is $w_{s}=\frac{b}{b + \mathbb{P}(s)}$, $\mathbb{P}(s)$ is the sentence density in a corpus $D$.
We empirically set $b$ as half of the average probability of all sentences, i.e., $b=\frac{1}{2T}\cdot \sum\limits_{s\in D}\mathbb{P}(s)$ with $T$ as the total number of sentences in corpus $D$.

\begin{algorithm}[htbp!]
\caption{LAWDR}
\label{algo}
\begin{algorithmic} 
\STATE \textbf{Input}: A corpus $D$ of documents $\{d: d\in D\}$ in the same language and each document $d$ consists of a set of sentences $\{s\}$. 
\STATE \textbf{Encode sentences}: Sentence embedding  $\mathbf{v}_{s}=\text{PretrainedModel}(s)$.
\STATE \textbf{Estimate bias}: $\tilde{\bf{u}}_{1}^{\text{lang}}, ..., \tilde{\bf{u}}_{m}^{\text{lang}} \leftarrow \text{SVD}(\{\mathbf{v}_{s}:s\in D\})$
\STATE \textbf{Debias}: $\tilde{\mathbf{v}}_{s} \leftarrow \mathbf{v}_{s} - \sum\limits_{i=1}^{m}\mathbf{v}^{T}_{s}\tilde{\bf{u}}^{\text{lang}}_{i}\tilde{\bf{u}}^{\text{lang}}_{i}$, $\forall s\in D$
\STATE \textbf{Estimate density}: $\{\mathbb{P}(s)\} \leftarrow \text{KDE}(\{\mathbf{v}_{s}:s\in D\})$
\STATE \textbf{Calculate weight}: $w_{s} \leftarrow \frac{b}{b + \mathbb{P}(s)}$, $\forall s\in D$, where $b=\frac{1}{2T}\cdot \sum\limits_{s\in D}\mathbb{P}(s)$ and $T$ is the total number of sentences 
\FOR{document $d \in D$}
\STATE \textbf{Encode documents}: $\mathbf{v}_{d} \leftarrow \sum\limits_{s\in d}w_{s}\tilde{\mathbf{v}}_{s}$
\ENDFOR
\RETURN document embeddings $\{\mathbf{v}_{d}:d\in D\}$
\end{algorithmic}
\end{algorithm}

We present LAWDR for document representation learning in Algorithm~\ref{algo}. Sentence representations are first generated by a pre-trained cross-lingual model (BERT, XLM or XLM-R in our study) using either special token (ST) representation or mean pooling (MP) over token embeddings. We then remove language bias from sentence embeddings. The document representations are derived from a weighted linear combination of debiased sentence representations, where sentence weights are their inverse density in the corpus.

\section{Document Similarity Metric}
\label{sec:doc_sim}

Document alignment is an important application of cross-lingual document representations, which also serves as an evaluation task for the semantic representation quality. The task is to measure the the semantic similarity of documents in a source and a target language. In this section, we study the geometry of the learned document representations and discuss the metric to measure document similarity from the embeddings.

\textbf{Geometry of document representations}. Cosine similarity is the most commonly used metric for language representations \cite{mikolov2013distributed,guo2018effective}. Given a document in a source language, we use the cosine similarity to find its nearest neighbors from a target language in the vector space. We observe that its parallel document is usually among its (top four) nearest neighbors but not the closest one. 

A possible explanation is that the document representation carries more than semantic information \cite{mu2018all}. Language is one kind of non-semantic signal, which has been removed in debiasing procedure. Domain, syntax and text style are examples of non-semantic signals that have been found in language representations \cite{mikolov2013distributed,nooralahzadeh2018evaluation}. The non-semantic component would change the local geometry of document embeddings. Hence cosine similarity may not accurately reflect the semantic closeness.

\textbf{Similarity metric}. A margin function has been recently proposed to evaluate multilingual sentence similarity which outperforms cosine function \cite{artetxe2018margin}. Suppose that there two corpora $D_{l}$ and $D_{l'}$ in language $l$ and $l'$ respectively. A margin function $\text{margin}(d,d')$ is defined for documents $d$ and  $d'$ from corpora $D_{l}$ and $D_{l'}$ respectively:
\begin{align}
\nonumber
&\text{margin}(\mathbf{v}_{d}, \mathbf{v}_{d'}) \\ 
&= \frac{\langle\mathbf{v}_{d},\mathbf{v}_{d'}\rangle}{
\sum\limits_{\mathbf{v}\in\text{knn}(\mathbf{v}_d, l')}\langle\mathbf{v}_{d}, \mathbf{v}\rangle
+{\sum\limits_{\mathbf{v}\in\text{knn}(\mathbf{v}_{d'},l)}\langle\mathbf{v}_{d'}, \mathbf{v}\rangle}},
\end{align}
where $\text{knn}(\mathbf{v}_d, l')$ is the embeddings of $k$ nearest neighbors in language $l'$ of document $d$ in terms of cosine similarity.

Given a source document, \citeauthor{artetxe2018margin} proposed to find parallel candidates using cosine similarity, and re-rank these candidates using margin function. The document with the maximal margin value is identified as the parallel document \cite{artetxe2018margin}.

\section{Experiments}
We evaluate the document representations on a semantic similarity task -- cross-lingual document alignment. It identifies parallel documents, and mines massive data for machine translation \cite{el2019massive}. 
\begin{table}[htbp!]
\centering
\resizebox{0.48\textwidth}{!}{
\begin{tabular}{c|ccc}
\hline
Dataset & \multicolumn{2}{c}{WMT-16} & WMT-19  \\ \hline
Language & En & Fr & 15 langs \\ \hline
Doc \# & 681,529 & 522,627 & 86,853  \\ \hline
Average length & 853 & 944 & 1108 \\ \hline
\end{tabular}}
\caption{Statistics of test data.}
\label{tab:data_stat}
\end{table}

\begin{table*}[htbp!]
\centering
\resizebox{1.0\textwidth}{!}{
\begin{tabular}{ccccccccc|cccccc}
\hline
Bias & Sent2Doc & Metric & \multicolumn{6}{c|}{No URL} & \multicolumn{6}{c}{With URL} \\ \hline
 &  &  & \multicolumn{2}{c}{BERT} & \multicolumn{2}{c}{XLM} & \multicolumn{2}{c|}{XLM-R} & \multicolumn{2}{c}{BERT} & \multicolumn{2}{c}{XLM} & \multicolumn{2}{c}{XLM-R} \\ \cline{4-15}
 &  &  & ST & MP & ST & MP & ST & MP & ST & MP & ST & MP & ST & MP \\ \hline
\multirow{4}{*}{Biased} & \multirow{2}{*}{Mean} & Cosine & 5.0 & 8.1 & 16.2 & 17.0 & 3.8 & 4.7 & 61.5 & 62.7 & 65.2 & 64.0 & 61.2 & 61.4 \\ 
 &  & Margin & 7.1 & 9.4 & 17.9 & 17.7 & 4.7 & 5.6 & 62.1 & 63.4 & 65.9 & 64.3 & 61.4 & 61.7 \\ \cline{2-15} 
 & \multirow{2}{*}{Weighted} & Cosine & 6.3 & 8.7 & 18.4 & 17.6 & 3.9 & 4.9 & 62.1 & 62.9 & 65.3 & 64.0 & 61.2 & 61.8 \\ 
 &  & Margin & 7.7 & 10.4 & 19.2 & 18.5 & 4.7 & 6.1 & 62.5 & 63.8 & 65.9 & 64.4 & 61.5 & 62.1 \\ \hline
\multirow{4}{*}{Debiased} & \multirow{2}{*}{Mean} & Cosine & 77.7 & 85.1 & 86.3 & 87.7 & 56.9 & 60.7 & 86.8 & 92.5 & 93.7 & 93.8 & 77.8 & 79.3 \\ 
 &  & Margin & \textbf{78.1} & 86.1 & \textbf{87.3} & 88.6 & \textbf{58.0} & 61.2 & \textbf{87.6} & 93.1 & 93.8 & 94.2 & \textbf{78.4} & 79.5 \\ \cline{2-15} 
 & \multirow{2}{*}{Weighted} & Cosine & 77.3 & 85.3 & 85.8 & 88.4 & 57.3 & 60.9 & 87.2 & 93.0 & 93.7 & 94.0 & 78.1 & \textbf{79.6} \\ 
 &  & Margin & 78.0 & \textbf{86.6} & 87.0 & \textbf{88.8} & 57.6 & \textbf{61.3} & 87.3 & \textbf{94.0} & \textbf{93.9} & \textbf{94.8} & 78.1 & 79.4 \\ \hline
\end{tabular}}
\caption{Document alignment recall (\%) of LAWDR with different pre-trained Models on WMT-16 data. (Without URL information, LASER baseline has a recall of $62.8\%$ when the documents are truncated to $512$ tokens, and a recall of $20.2\%$ when encoding full documents. With URL information, LASER achieves recalls of $84.7\%$ and $69.5\%$ when encoding truncated and full documents respectively.)}
\label{tab:wmt16}
\end{table*}

\noindent\textbf{Dataset}. We used two publicly available datasets from WMT-16 and WMT-19 shared tasks for experiments on cross-lingual document alignment. WMT-16 dataset\footnote{\url{http://www.statmt.org/wmt16/bilingual-task.html}} is prepared by the shared task on bilingual alignment between English and French documents from 203 websites \cite{buck2016findings}. The other dataset consisting of new commentary is provided in WMT-19 task \footnote{\url{http://data.statmt.org/news-commentary/v14/}}, covering 15 languages\footnote{Arabic, Czech, German, English, Spanish, French, Hindi, Indonesian, Italian, Japanese, Kazakh, Dutch, Portuguese, Russian and Chinese} and 99 language pairs.
Table~\ref{tab:data_stat} shows the statistics of datasets used in our experiments including the number and the average length of their documents.

\noindent\textbf{Evaluation}. Following the evaluation of WMT-16 shared task, we use \emph{Recall} as the metric of document alignment. Given two sets of documents in two languages, we sort cross-lingual document pairs in decreasing order of their semantic similarity estimated by cosine and margin metrics. Recall measures the percentage of aligned document pairs that can be correctly identified by models. The higher the recall is, the better a model performs.

\subsection{Results}

\noindent\textbf{Setup}. Documents are first segmented into sentences, and then we obtain the sentence embeddings from pre-trained models. Our experiments cover three recent cross-lingual models -- multilingual BERT \cite{devlin2019bert}, XLM \cite{lample2019cross} and XLM-R \cite{conneau2019unsupervised}. Algorithm \ref{algo} is applied to derive document representations from their sentence embeddings. We compare margin and cosine functions as measures of document similarity. 

\noindent\textbf{Baselines}. We include state-of-the-art sentence encoder -- LASER\footnote{\url{https://github.com/facebookresearch/LASER}} -- as a baseline for document representation. It achieves strong performance in aligning cross-lingual sentences~\cite{chaudhary-EtAl:2019:WMT}. Since it is an LSTM-based model without length limits on input texts, we apply it to document representation learning in our experiments. To understand how its representation quality changes with input length, we report its alignment performance of encoding whole documents and encoding only the first $512$ tokens respectively. 

The average of sentence embedding has been shown to be a strong approach to derive document representation \cite{guo2019hierarchical}. We include it as another strong baseline in our experiments, with which we can analyze the gains brought by sentence weighting. 

\noindent\textbf{WMT16 results}. The state-of-the-art system in this dataset used machine translation to compute lexical overlap of documents \cite{gomes2016first}. It achieved recalls of $85.8\%$ and $95.0\%$ without and with URL information respectively. 

Table \ref{tab:wmt16} reports the recall of LAWDR combined with three pre-trained models on WMT-16 dataset. For completeness, it includes the results w/o URL information in the columns ``no URL'' and ``with URL'' respectively. To use the URL information, We simply use the URL matching script provided by the WMT-16 shared task to align some of the documents \cite{buck2016findings}. Then we score the similarity of the remaining document pairs with the document representations.   

As shown in Table \ref{tab:wmt16}, LASER representation suffers from quality degradation as the text length increases. The best-performing approach is the combination of the document embedding as a weighted average of debiased XLM+MP embedding and margin as similarity metric.
It achieves a recall of $88.8\%$ without URL and a recall of $94.8\%$ with URL information. This is comparable to the state-of-the-art model built upon machine translation. Moreover, it outperforms LASER by a large margin no matter whether URL matching is used.

\begin{table*}[htbp!]
\centering
\begin{tabular}{ccccccccc}
\hline
Bias & Sent2Doc & Metric & \multicolumn{2}{c}{BERT} & \multicolumn{2}{c}{XLM} & \multicolumn{2}{c}{XLM-R} \\ \hline
 &  &  & ST & MP & ST & MP & ST & MP \\ \hline
\multirow{4}{*}{Biased} & \multirow{2}{*}{\begin{tabular}[c]{@{}c@{}}Mean\end{tabular}} & Cosine & 1.0 & 4.6 & 6.8 & 8.1 & 0.1 & 0.1 \\ 
 &  & Margin & 1.5 & 5.8 & 13.0 & 15.6 & 0.1 & 0.1  \\ \cline{2-9} 
 & \multirow{2}{*}{\begin{tabular}[c]{@{}c@{}}Weighted\end{tabular}} & Cosine & 1.3 & 5.2 & 7.2 & 8.5 & 0.1 & 0.1 \\ 
 &  & Margin & 1.8 & 6.4 & 13.3 & 16.1 & 0.1 & 0.1 \\ \hline
\multirow{4}{*}{Debiased} & \multirow{2}{*}{\begin{tabular}[c]{@{}c@{}}Mean\end{tabular}} & Cosine & 57.6 & 76.1 & 85.1 & 89.8 & 23.9 & 42.2 \\ 
 &  & Margin & 60.6 & 79.0 & 87.1 & 90.7 & 25.7 & 47.0 \\ \cline{2-9} 
 & \multirow{2}{*}{\begin{tabular}[c]{@{}c@{}}Weighted\end{tabular}} & Cosine & 57.8 & 75.9 & 85.5 & 90.0 & 25.5 & 42.8 \\ 
 &  & Margin & \textbf{61.0} & \textbf{79.5} & \textbf{87.6} & \textbf{91.0} & \textbf{26.0} & \textbf{47.6} \\ \hline
\end{tabular}
\caption{Document alignment recall (\%) of LAWDR with different pre-trained models on WMT-19 data. (The  LASER baseline achieves a recall of $68.0\%$ when encoding the first $512$ tokens of the documents and a recall of $3.0\%$ when encoding full documents.)}
\label{tab:wmt19}
\end{table*}

\noindent\textbf{WMT19 results}. The results on WMT-19 dataset are reported in Table \ref{tab:wmt19}. 
We again observe that LASER's representation quality get severely worse as its input texts become longer. With a recall of $91.0\%$, our best model is again the combination of weighted document representation from debiased XLM MP embedding and margin metric. 

\subsection{Analysis}
We now analyze the effect of each step in LAWDR algorithm on document alignment performance.

\textbf{MP v. ST}. For all three pre-trained models, MP consistently provides better sentence embeddings than ST sentence embedding.

\textbf{Debiasing}. Debiasing is surprisingly important to the embedding quality and alignment performance.  Consider the setting that weighted document representation is derived from MP sentence embedding and margin is the metric. On WMT16 without URL, debiasing brings a gain of $76.2\%$ for BERT, $70.3\%$ for XLM and $55.2\%$ for XLM-R. On WMT19, debiasing achieves a gain of $73.1\%$ for BERT, $74.9\%$ for XLM and $47.5\%$ for XLM-R.

\textbf{Sentence weighting}. Weighted document representation shows its advantage over mean representations. Consider the combination of XLM, MP and margin metric on WMT16. With original biased sentence embedding, weighted representation improves the alignment recall by $0.8\%$ in comparison with mean representation. The improvement is $0.2\%$ for debiased sentence embeddings. Similar gains could be observed on WMT19 data.

\textbf{Margin v. Cosine}. As for the similarity metric, margin function outperforms cosine function in most of the cases. Consider the weighted document representation derived from XLM MP. On WMT16 dataset, the gain of margin metric over cosine is $0.9\%$ on biased embedding, and $0.4\%$ on debiased embedding. On WMT19, the gain of margin is $7.6\%$ with biased embedding, and $1.0\%$ with debiased embedding.

\section{Discussion}
In our study, the pre-trained models are applied to generate representations for single sentences. It is natural to consider encoding multiple sentence so that these model could capture the  inter-sentence information \cite{ethayarajh2019contextual}. Now we take as many sentences as possible under the length limit of the models, and concatenate them as one input sequence. Again we use MP for sentence embeddings, LAWDR for weighted document representations and margin function for the similarity metric.  

\begin{table}[htbp!]
\centering
\resizebox{0.48\textwidth}{!}{
\begin{tabular}{ccccc}
\hline
\multicolumn{1}{c}{Input} & Bias & BERT & XLM & XLM-R \\ \hline
\multirow{2}{*}{\begin{tabular}[c]{@{}c@{}}Single\\ Sent\end{tabular}} & Biased & 6.4 & 16.1 & 0.1 \\ 
 & Debiased & \textbf{79.5} & \textbf{91.0} & \textbf{47.6} \\ \hline
\multirow{2}{*}{\begin{tabular}[c]{@{}c@{}}Multi\\ Sent\end{tabular}} & Biased & 21.0 & 20.7 & 1.9 \\ 
 & Debiased & \textbf{86.3} & \textbf{88.4} & \textbf{74.2} \\ \hline
\end{tabular}}
\caption{Recall (\%) of different inputs on WMT19}
\label{tab:multi_sent_wmt19}
\end{table}

Table~\ref{tab:multi_sent_wmt19} compares WMT19 recalls with single and multiple sentences as model inputs. With biased embedding, multi-sentence inputs outperform single-sentence inputs for all models. With debiased embedding, multi-sentence inputs yield gains of $6.8\%$ for BERT and $26.6\%$ XLM-R, while falls behind single-sentence inputs by $2.6\%$ for XLM. The complete results of all models on both datasets are provided in the supplementary material.

\section{Related Work}
\noindent\textbf{Multilingual sentence representation}. Learning multilingual embeddings is appealing due to their universal representations across languages. Represented in the same embedding space, the semantics of languages can be compared directly regardless of their languages in a scalable and efficient way \cite{conneau2018xnli}. Moreover, multilingual representations enable model transfer between languages especially from high-resource to low-resource languages \cite{ruder2017survey}. 

There has been a long research venue in multilingual sentence representations learning. Recently NLP has seen a surge in the study of multilingual pre-trained models. 
One of the strong models is Language-Agnostic SEntence Representations (LASER) \cite{artetxe2018massively}, which is built on LSTM network. Its representation quality degrades with the increase of text length.
Other cross-lingual models are based on Transformer \cite{vaswani2017attention} such as BERT \cite{devlin2019bert}, XLM \cite{lample2019cross} and  XLM-RoBERTa \cite{conneau2019unsupervised}. These models are not able to encode long documents due to strict limits on inputs.



\noindent\textbf{Document embedding}. Monolingual document representations has been extensively studied. Document vector is trained to predict component words \cite{gupta2016doc2sent2vec}. Document embeddings are also learned from word embeddings \cite{chen2016efficient}. Recent works propose end-to-end hierarchical models to learn document embeddings on labeled data \cite{yang2016hierarchical,miculicich2018document}. 

Despite a large body of research on monolingual document representations, there have been very few works in cross-lingual setting. The most recent work designs hierarchical multilingual document encoder (HiDE), which is trained on a large corpus of parallel documents \cite{guo2019hierarchical}. It only experiments with English-French and English-Spanish language pairs. Considering the large cost of training the model on other languages, we do not include it in our experiments.

\section{Conclusion}
In this paper, we proposed LAWDR to learn language-agnostic document representation on top of pre-trained cross-lingual models. We first apply debiasing to mitigate language signals and derive language-agnostic sentence embeddings. The debiased sentence embeddings are then weighted with their inverse density as document representations. We further reveal that margin function is a more accurate measure of document similarity than cosine similarity with empirical evaluation. The learned document embeddings have achieved state-of-the-art performance on cross-lingual document alignment.



\bibliographystyle{acl_natbib}
\bibliography{acl2021}
\newpage
\section{Supplemental Material}
\label{sec:supplemental}

\subsection{Language Bias in Sentence Embeddings}

\begin{figure*}[h]
\centering
\begin{minipage}{0.32\textwidth}
\centerline{\includegraphics[width=\linewidth]{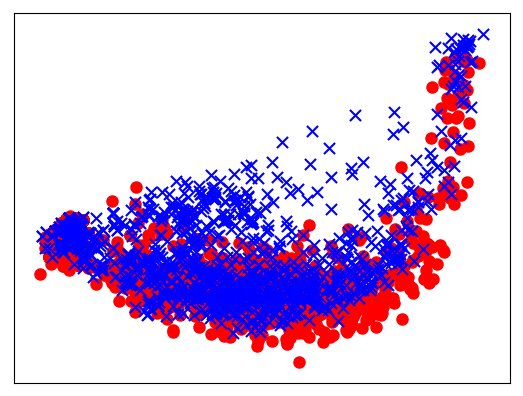}}
\centerline{\small{(a) BERT embedding.}}
\label{fig:laser_cls_bias}
\end{minipage}
\begin{minipage}[c]{0.32\textwidth}
\centerline{\includegraphics[width=\linewidth]{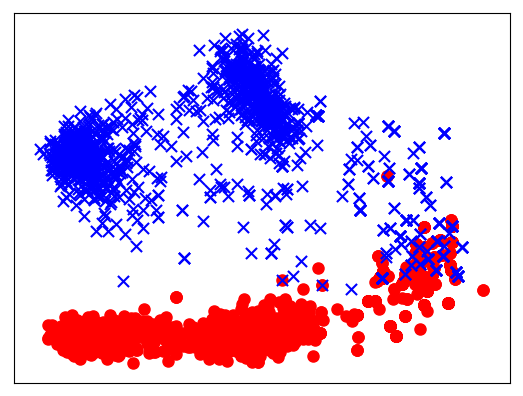}}
\centerline{\small{(b) XLM embedding.}}
\label{fig:bert_cls_bias}
\end{minipage}
\begin{minipage}[c]{0.32\textwidth}
\centerline{\includegraphics[width=\linewidth]{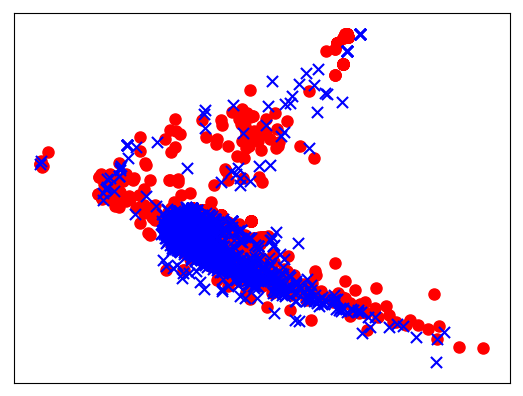}}
\centerline{\small{(c) XLM-R embedding.}}
\label{fig:xlm_cls_bias}
\end{minipage}
\caption{Visualization of sentence embeddings by pre-trained BERT and XLM models. English sentences are marked as red and French sentences are marked as blue.}
\label{fig:app_visual}
\end{figure*}

We derive sentence embeddings with the special token (ST) embeddings. Figure~\ref{fig:app_visual} visualizes the embeddings of English and French sentences from parallel eu2007.de documents. We note that sentences in different languages are separated in the vector space. It suggests that pre-trained sentence embeddings are biased towards languages.

\noindent\textbf{Debiasing sentence embeddings}. We collect embeddings of sentences in the same language, and sort the domain components in decreasing order of the captured variance ratio using SVD. The top $m$ components are identified as the language components. To decide the value of $m$, we find the smallest number of dominant components so that the language classification accuracy of debiased embeddings is lower than $55\%$. This resulted in $m=64$ for BERT embeddings, $m=32$ for XLM embeddings and $m=256$ for XLM-R embeddings.

\begin{table}[htbp!]
\centering
\begin{tabular}{cccc}
\hline
Embedding & BERT & XLM & XLM-R \\ \hline
$\{\mathbf{v}_{s}\}_{s}$ & 95.3 & 97.3 & 93.1 \\ 
$\{\mathbf{v}_{s}^{\text{lang}}\}_{s}$ & 99.3 & 100.0 & 96.7 \\ 
$\{\mathbf{v}_{s}^{\text{sem}}\}_{s}$ & 52.0 & 55.5 & 52.0 \\ \hline
\end{tabular}
\caption{Accuracy of language classification with ST sentence embeddings.}
\label{tab:app_lang_clf}
\end{table}

\subsection{Experiment}
\label{sec:app_exp}

\noindent\textbf{Pre-trained cross-lingual models}. 
BERT model is pre-trained on $102$ languages with the largest Wikipedia corpus, and its training objective is masked language modeling as well as next sentence prediction. XLM and XLM-R models are pre-trained on 100 languages with masked language modeling. We use the implementation of HugginFace to obtain token embeddings from these pre-trained models \cite{Wolf2019HuggingFacesTS}.

\noindent\textbf{Hyperparameters}. We analyze the components in sentence embeddings with SVD, and debias sentence embeddings when the dominant components are removed as language bias. To assign weights of sentences, we estimate their density using kernel density estimation with Tophat as the kernel. It is known that the complexity of kernel density estimation grows exponentially with the dimension \cite{scott2015multivariate}. To make the density estimation more computationally efficient, we first reduce the embedding dimension to $d'=16$ with principal component analysis (PCA), and apply density estimation to reduced embeddings. The bandwidth in density estimation is selected using $5$-fold cross validation on each dataset. 

For sentence density estimation, we apply kernel density estimator to sentence embeddings, and use Tophat kernel. Since the complexity of density estimation grows with the dimension of embeddings, we project sentence embeddings to a $16$-dimension space using principal component analysis. The embeddings of reduced dimension are used as the input to density estimator.

\begin{table*}[htbp!]
\centering
\resizebox{\textwidth}{!}{
\begin{tabular}{cccccccccc|c|c|c}
\hline
\multicolumn{13}{c}{\textbf{No URL}} \\ \hline
 &  &  &  & \multicolumn{2}{c}{BERT} & \multicolumn{2}{c}{XLM} & \multicolumn{2}{c}{XLM-R} & \multicolumn{2}{c}{LASER} & SoA \\ \hline
 &  &  &  & ST & MP & ST & MP & ST & MP & 512 tokens & all tokens & - \\ \hline
\multirow{8}{*}{\begin{tabular}[c]{@{}c@{}}Single\\ Sent\end{tabular}} & \multirow{4}{*}{Biased} & \multirow{2}{*}{Mean} & Cosine & 5.0 & 8.1 & 16.2 & 17.0 & 3.8 & 4.7 & \multirow{16}{*}{62.8} & \multirow{16}{*}{20.2} & \multirow{16}{*}{85.8} \\ 
 &  &  & Margin & 7.1 & 9.4 & 17.9 & 17.7 & 4.7 & 5.6 &  &  &  \\ \cline{3-10}
 &  & \multirow{2}{*}{Weighted} & Cosine & 6.3 & 8.7 & 18.4 & 17.6 & 3.9 & 4.9 &  &  &  \\ 
 &  &  & Margin & 7.7 & 10.4 & 19.2 & 18.5 & 4.7 & 6.1 &  &  &  \\ \cline{2-10}
 & \multirow{4}{*}{Debiased} & \multirow{2}{*}{Mean} & Cosine & 77.7 & 85.1 & 86.3 & 87.7 & 56.9 & 60.7 &  &  &  \\ 
 &  &  & Margin & \textbf{78.1} & 86.1 & \textbf{87.3} & 88.6 & \textbf{58.0} & 61.2 &  &  &  \\ \cline{3-10}
 &  & \multirow{2}{*}{Weighted} & Cosine & 77.3 & 85.3 & 85.8 & 88.4 & 57.3 & 60.9 &  &  &  \\ 
 &  &  & Margin & 78.0 & \textbf{86.6} & 87.0 & \textbf{88.8} & 57.6 & \textbf{61.3} &  &  &  \\ \cline{1-10}
\multirow{8}{*}{\begin{tabular}[c]{@{}c@{}}Multi\\ Sent\end{tabular}} & \multirow{4}{*}{Biased} & \multirow{2}{*}{Mean} & Cosine & 8.7 & 25.7 & 18.3 & 25.1 & 3.0 & 4.7 &  &  &  \\ 
 &  &  & Margin & 8.7 & 28.2 & 20.5 & 26.9 & 3.3 & 5.5 &  &  &  \\ \cline{3-10}
 &  & \multirow{2}{*}{Weighted} & Cosine & 9.1 & 29.2 & 19.1 & 25.9 & 2.9 & 6.6 &  &  &  \\ 
 &  &  & Margin & 9.1 & 30.6 & 21.2 & 27.4 & 3.2 & 7.2 &  &  &  \\ \cline{2-10}
 & \multirow{4}{*}{Debiased} & \multirow{2}{*}{Mean} & Cosine & 68.4 & 84.9 & 82.1 & 85.9 & 49.8 & 65.9 &  &  &  \\ 
 &  &  & Margin & 68.7 & 85.5 & 83.1 & 86.6 & 49.9 & 66.0 &  &  &  \\ \cline{3-10}
 &  & \multirow{2}{*}{Weighted} & Cosine & 68.1 & 85.0 & 82.3 & 87.6 & 49.9 & 65.9 &  &  &  \\ 
 &  &  & Margin & \textbf{68.7} & \textbf{85.6} & \textbf{83.1} & \textbf{88.3} & \textbf{50.2} & \textbf{66.1} &  &  &  \\ \hline
\multicolumn{13}{c}{\textbf{With URL}} \\ \hline
 &  &  &  & \multicolumn{2}{c}{BERT} & \multicolumn{2}{c}{XLM} & \multicolumn{2}{c|}{XLM-R} & \multicolumn{2}{c|}{LASER} & SoA \\ \hline
 &  &  &  & ST & MP & ST & MP & ST & MP & 512 tokens & all tokens & - \\ \hline
\multirow{8}{*}{\begin{tabular}[c]{@{}c@{}}Single\\ Sent\end{tabular}} & \multirow{4}{*}{Biased} & \multirow{2}{*}{Mean} & Cosine & 61.5 & 62.7 & 65.2 & 64.0 & 61.2 & 61.4 & \multirow{16}{*}{84.7} & \multirow{16}{*}{69.5} & \multirow{16}{*}{95.0} \\ 
 &  &  & margin & 62.1 & 63.4 & 65.9 & 64.3 & 61.4 & 61.7 &  &  &  \\ \cline{3-10}
 &  & \multirow{2}{*}{Weighted} & Cosine & 62.1 & 62.9 & 65.3 & 64.0 & 61.2 & 61.8 &  &  &  \\ 
 &  &  & Margin & 62.5 & 63.8 & 65.9 & 64.4 & 61.5 & 62.1 &  &  &  \\ \cline{2-10}
 & \multirow{4}{*}{} & \multirow{2}{*}{Mean} & Cosine & 86.8 & 92.5 & 93.7 & 93.8 & 77.8 & 79.3 &  &  &  \\ \cline{4-10}
 &  &  & Margin & \textbf{87.6} & 93.1 & 93.8 & 94.2 & \textbf{78.4} & \textbf{79.5} &  &  &  \\ \cline{3-10}
 &  & \multirow{2}{*}{Weighted} & Cosine & 87.2 & 93.0 & 93.7 & 94.0 & 78.1 & 79.6 &  &  &  \\ 
 &  &  & Margin & 87.3 & \textbf{94.0} & \textbf{93.9} & \textbf{94.8} & 78.1 & 79.4 &  &  &  \\ \cline{1-10}
\multirow{8}{*}{\begin{tabular}[c]{@{}c@{}}Multi\\ Sent\end{tabular}} & \multirow{4}{*}{Biased} & \multirow{2}{*}{Mean} & Cosine & 62.3 & 66.4 & 65.0 & 66.7 & 60.7 & 61.5 &  &  &  \\ 
 &  &  & Margin & 62.4 & 67.6 & 65.9 & 67.6 & 61.0 & 61.7 &  &  &  \\ \cline{3-10}
 &  & \multirow{2}{*}{Weighted} & Cosine & 62.3 & 67.9 & 65.2 & 67.0 & 60.9 & 62.1 &  &  &  \\ 
 &  &  & Margin & 62.7 & 68.3 & 66.1 & 67.7 & 61.0 & 62.1 &  &  &  \\ \cline{2-10}
 & \multirow{4}{*}{Debiased} & \multirow{2}{*}{Mean} & Cosine & 83.2 & 92.9 & 90.6 & 92.8 & 75.4 & 82.8 &  &  &  \\ 
 &  &  & Margin & 83.6 & 93.1 & 91.3 & 93.2 & 75.3 & 82.9 &  &  &  \\ \cline{3-10}
 &  & \multirow{2}{*}{Weighted} & Cosine & 83.3 & 93.0 & 90.9 & 93.8 & 75.4 & 82.8 &  &  &  \\ 
 &  &  & Margin & 83.3 & \textbf{93.2} & \textbf{91.3} & 94.5 & \textbf{75.6} & 83.1 &  &  &  \\ \hline
\end{tabular}}
\caption{Document alignment recall (\%) of LAWDR with different pre-trained models on WMT-16 data.}
\label{tab:app_wmt16}
\end{table*}

\noindent\textbf{WMT16 results}. Table~\ref{tab:app_wmt16} presents the full results of bilingual document alignment on WMT-16 dataset.

\begin{table*}[htbp!]
\centering
\begin{tabular}{cccccccccc|c|c}
\hline
\multicolumn{1}{c}{Input} & Bias & Sent2Doc & Metric & \multicolumn{2}{c}{BERT} & \multicolumn{2}{c}{XLM} & \multicolumn{2}{c|}{XLM-R} & \multicolumn{2}{c}{LASER} \\ \hline
\multicolumn{1}{c}{} &  &  &  & ST & MP & ST & MP & ST & MP & 512 tokens & all tokens \\ \hline
\multirow{8}{*}{\begin{tabular}[c]{@{}c@{}}Single\\ Sent\end{tabular}} & \multirow{4}{*}{Biased} & \multirow{2}{*}{Mean} & Cosine & 1.0 & 4.6 & 6.8 & 8.1 & 0.1 & 0.1 & \multirow{16}{*}{68.0} & \multirow{16}{*}{3.0} \\ 
 &  &  & Margin & 1.5 & 5.8 & 13.0 & 15.6 & 0.1 & 0.1 &  &  \\ \cline{3-10}
 &  &  \multirow{2}{*}{Weighted} & Cosine & 1.3 & 5.2 & 7.2 & 8.5 & 0.1 & 0.1 &  &  \\ 
 &  &  & Margin & 1.8 & 6.4 & 13.3 & 16.1 & 0.1 & 0.1 &  &  \\ \cline{2-10}
 & \multirow{4}{*}{Debiased} & \multirow{2}{*}{Mean} & Cosine & 57.6 & 76.1 & 85.1 & 89.8 & 23.9 & 42.2 &  &  \\ 
 &  &  & Margin & 60.6 & 79.0 & 87.1 & 90.7 & 25.7 & 47.0 &  &  \\ \cline{3-10}
 &  &  \multirow{2}{*}{Weighted}  & Cosine & 57.8 & 75.9 & 85.5 & 90.0 & 25.5 & 42.8 &  &  \\ 
 &  &  & Margin & \textbf{61.0} & \textbf{79.5} & \textbf{87.6} & \textbf{91.0} & \textbf{26.0} & \textbf{47.6} &  &  \\ \cline{1-10}
\multirow{8}{*}{\begin{tabular}[c]{@{}c@{}}Multi\\ Sent\end{tabular}} & \multirow{4}{*}{Biased} & \multirow{2}{*}{Mean} & Cosine & 7.4 & 12.3 & 9.5 & 12.3 & 0.2 & 0.6 &  &  \\ 
 &  &  & Margin & 8.9 & 17.0 & 16.6 & 20.9 & 0.2 & 1.2 &  &  \\ \cline{3-10}
 &  & \multirow{2}{*}{Weighted} & Cosine & 8.1 & 14.8 & 9.5 & 12.2 & 0.2 & 1.6 &  &  \\ 
 &  &  & Margin & 10.0 & 21.0 & 16.5 & 20.7 & 0.2 & 1.9 &  &  \\ \cline{2-10}
 & \multirow{4}{*}{Debiased} & \multirow{2}{*}{Mean} & Cosine & 59.6 & 84.3 & 78.5 & 86.9 & 41.3 & 72.8 &  &  \\ 
 &  &  & Margin & 62.7 & 85.9 & 80.7 & 88.0 & 42.4 & 74.1 &  &  \\ \cline{3-10}
 &  & \multirow{2}{*}{Weighted} & Cosine & 59.7 & 85.7 & 78.5 & 87.3 & 41.3 & 73.2 &  &  \\ 
 &  &  & Margin & \textbf{62.8} & \textbf{86.3} & \textbf{80.8} & \textbf{88.4} & \textbf{42.4} & \textbf{74.2} &  &  \\ \hline
\end{tabular}
\caption{Document alignment recall (\%) of LAWDR with different pre-trained models on WMT-19 data.}
\label{tab:app_wmt19}
\end{table*}

\noindent\textbf{WMT19 results}. Table~\ref{tab:app_wmt19} presents the full results on WMT-19 multilingual document alignment. 


\end{document}